\documentclass[letter, 10pt, conference]{ieeeconf} 
\IEEEoverridecommandlockouts                              % This command is only
\usepackage[OT1]{fontenc} 

\usepackage{cite}
\usepackage{float}
\usepackage{amsmath}
\usepackage{caption,subcaption}%
\usepackage{amsfonts}
\usepackage{graphicx}

\usepackage{booktabs}
\usepackage{placeins}
\usepackage{xr}
\usepackage{bbding}
\usepackage{array}

\usepackage{relsize}
\usepackage{graphicx}
\usepackage{esint}
\usepackage[normalem]{ulem}
\usepackage{arydshln}
\usepackage{eurosym}
\usepackage{caption}
\usepackage{color}
\usepackage{amsmath,amssymb}
\usepackage{subcaption}  % \begin{subfigure}...\end{subfigure} within figure
\usepackage{multirow}
\usepackage{textcomp}
\usepackage{tabularx}
\usepackage{placeins}
\usepackage{lscape}
\usepackage{comment}
\usepackage{mathrsfs}
\usepackage{algorithm2e}
\usepackage{mathtools}

\title{\LARGE \bf
SAPI: Surroundings-Aware Vehicle Trajectory Prediction at Intersections
}

\author{Ethan Zhang$^{1}$,
Hao Xiao$^{2}$, 
Yiqian Gan$^{2}$ and
Lei Wang$^{2}$
\thanks{The work was done at TuSimple Inc. when E. Zhang was a research intern.
$^{1}$E. Zhang is with
        University of Michigan, Ann Arbor, MI, USA, 48105
        {\tt\small shuruiz@umich.edu}.
        $^{2}$H. Xiao, Y. Gan, and L. Wang are with 
        Tusimple Inc., San Diego, CA, USA, 92122
        }
}
\begin{document}

\maketitle
\thispagestyle{empty}
\pagestyle{empty}

%%%%%%%%%%%%%%%%%%%%%%%%%%%%%%%%%%%%%%%%%%%%%%%%%%%%%%%%%%%%%%%%%%%%%%%%%%%%%%%%
\begin{abstract}
In this work we propose a deep learning model, i.e., SAPI, to predict vehicle trajectories at intersections.  
SAPI uses an abstract way to represent and encode surrounding environment by utilizing information from real-time map, right-of-way, and surrounding traffic. The proposed model consists of two convolutional network (CNN) and recurrent neural network (RNN)-based encoders and one decoder. A refiner is proposed to conduct a look-back operation inside the model, in order to make full use of raw history trajectory information. We evaluate SAPI on a proprietary dataset collected in real-world intersections through autonomous vehicles. It is demonstrated that SAPI shows promising performance when predicting vehicle trajectories at intersection, and outperforms benchmark methods. The average displacement error(ADE) and final displacement error(FDE) for 6-second prediction are 1.84m and 4.32m respectively. We also show that the proposed model can accurately predict vehicle trajectories in different scenarios. 

\end{abstract}

\section{Introduction}
\noindent The thriving autonomous vehicle(AV) technology is gaining more and more public attention. %AVs are expected to operate without or with little human intervention, and are envisioned to revolutionize existing transportation systems \cite{fagnant2015preparing}. 
Video analytics with data captured by multiple cameras on the autonomous vehicle have been of high significance for many applications, e.g., the estimation of traffic flow characteristics, multi-camera tracking \cite{tang2018single}, vehicle/person re-identification \cite{xiao2018group,lin2019group}, and motion prediction \cite{he2020ust, deo2018convolutional, ding2019online, roy2019vehicle, gan2023mgtr}.

To safely interact with surrounding environment, AVs are supposed to demonstrate remarkable performance to accurately predict trajectories of surrounding vehicles, so as to avoid potential collisions and risks \cite{woo2017lane,ammoun2009real,park2018sequence}. In addition to safety concerns, accurate trajectory prediction on surrounding vehicles also enhances the performance of model predictive control (MPC) for autonomous vehicles \cite{wang2019trajectory,ji2016path}. These benefits making vehicle trajectory prediction an essential task when we are entering the era of next generation transportation systems \cite{fagnant2015preparing}. Although vital, vehicle trajectory prediction, especially prediction for long look-ahead time,  can be challenging due to the rapidly changing surrounding environment of vehicles and complex nature of driving behavior, which can be greatly affected by a number of factors, including driver's personality, road curvature, weather, traffic rules, etc.\cite{kim2017probabilistic}. 
 
This work focuses on predicting vehicle trajectories under one of the most challenging driving environment--intersections, where 50\% of total crashes happen \cite{usdot_2018}. In this paper we propose a deep learning-based approach, i.e., SAPI, to predict vehicle trajectories at intersections within long look-ahead time. Through a proposed environment representation strategy, SAPI incorporates real-time map, right-of-way and surrounding vehicle dynamics information in an abstract way. History trajectory of target vehicle is also used as one of the model inputs. Two encoders are introduced based on CNN and RNN in SAPI to learn patterns from surrounding environment and trajectory features separately. A refiner is proposed to refine the learned patterns by bridging outputted context patterns of encoder and raw history trajectory, conducting a look-back operation to further make the use of history information. A decoder is then used in the work to decode learned patterns and generate predicted future trajectories.  

Unlike existing work that exhaustively enumerating factors that may affect driving behavior \cite{cui2019multimodal, hong2019rules}, or modeling a certain type of factors \cite{deo2018convolutional,messaoud2019non}, e.g., vehicle interactions, SAPI separates information into two main types, i.e., surrounding environment and status of the target vehicle, and use an abstract environment representation strategy to encode surrounding environment. We show that the proposed model gains promising performance on a proprietary dataset collected by autonomous vehicles at variety of real-world intersections in Arizona, USA, and outperforms benchmark methods. When predicting future 6-second vehicle trajectory, ADE of the the proposed model is 1.84m and FDE is 4.32m. Considering the average vehicle length, which is around 4.2m, we claim that the proposed model demonstrates good performance. We also show that when the environment representation is partially unavailable, the model performance can be impaired. The proposed model demonstrates good performance in critic driving scenarios. For example, even though the traffic light information is not included in training data, the model can still anticipate driving intention and get accurate predictions by learning from surrounding traffic. 
The proposed model is light-weighted. Even though the number of parameters are much fewer than the benchmark method, it still shows significantly better performance in terms of ADE and FDE along the prediction horizon, which can be desirable for real-world applications.

\section{Related Work}
\noindent Current work focusing on vehicle trajectory prediction can be summarized into two categories, i.e., physics-based models and learning-based models. Physics-based models usually formulate the problem from physics' point of view, for example, developing robust motion models, etc., while learning-based models focus more on learning patterns of history trajectory of a target vehicle and its surrounding environment. In this section we review existing literature on vehicle trajectory prediction, with the emphasis on learning-based models, which is the category the proposed work belongs to. 

Physics-based models are mainly developed through manipulating motion models or state filtering strategies. A. Houenou et al. proposed a vehicle trajectory prediction method based on motion model and maneuver recognition \cite{houenou2013vehicle}. It combined trajectories predicted by constant yaw rate and acceleration motion model \cite{schubert2008comparison}, and trajectories predicted by maneuver recognition. By taking benefits of both models, it showed better prediction accuracy for both short-term and long-term predictions. S. G. Yi et al. proposed a vehicle trajectory
prediction algorithm for adaptive cruise control (ACC) purpose \cite{yi2015vehicle}. It used the yaw rate of the target vehicle and road curvature information, and then formulated future vehicle dynamics through a transfer function of integration on velocity and acceleration. C. Barrios et al. set up a dead reckoning system and conducted vehicle trajectory prediction based on Kalman filter \cite{barrios2015trajectory}, where vehicle dynamics were computed through constant velocity and acceleration models. Physics-based models are generally easier to set up, however, their performance may largely affected by prediction horizon. Complex nature of driving behaviors and challenges brought by dynamic driving environment can make physics-based models only suitable for short-term trajectory prediction \cite{xie2017vehicle}.

With advancements in machine learning and artificial intelligence in past decades, learning-based methods become the mainstream with respect to the topic\cite{he2020ust, deo2018convolutional, ding2019online, roy2019vehicle, gan2023mgtr}. N. Deo et al. proposed a deep learning model for vehicle trajectory prediction based on convolutional social pooling \cite{deo2018convolutional}. It used a long short-term memory (LSTM) encoder-decoder model along with convolutional social pooling to learn inter-dependencies between vehicles. It showed that the model demonstrated good performance on public datasets. However, as indicated by authors, it used purely vehicle tracks, which negatively affected prediction performance. 
% ding2019online, pay attention to benchmark writing
W. Ding et al. proposed a two-level vehicle trajectory prediction framework for urban driving scenarios \cite{ding2019online}. It used a LSTM network to anticipate vehicle's driving policy, i.e., driving forward, yield, turning left, or turning right, and then generated vehicle trajectories through optimization by minimizing the cost of driving context. It demonstrated good flexibility under different driving scenarios. However, it assumed deterministic reasoning based on one selected policy, which harmed its prediction accuracy. Based on \cite{gupta2018social}, D. Roy et al. proposed a vehicle trajectory prediction method based on generative adversarial networks (GAN) \cite{roy2019vehicle}. It modeled vehicle interactions in a social context, and generated the most acceptable future trajectory. It indicated that the proposed method was able to predict different traffic maneuvers like overtaking, merging, etc..
H. Cui et al. proposed a deep CNN-based model to predict vehicle trajectories \cite{cui2019multimodal}. %It utilized a multi-model setting and predicted several future trajectories as well as their probabilities. 
It used detailed surrounding information and compressed them into raster images. After that the information was encoded and learned by a constructed single deep CNN, and decoded into future trajectories. Through a multi-model strategy, authors demonstrated that the method showed good performance in terms of left-turn, straight and right-turn tasks. However, due to the model setting, the prediction results can be significantly affected by the number of modes, requiring the model to be carefully calibrated, and large computational resources were needed to train the model.  
J. Hong et al. proposed a convolutional model to predict driving behavior with semantic interactions \cite{hong2019rules}. It represented environment and semantic scene context into occupancy grid maps with 20 channels. Using a encoder-decoder pair, future trajectories were predicted along with probability distributions. It indicated that the model outperformed linear and Gaussian mixture models, and can create diverse samples for planning applications. However, exhaustively introduce as much information as possible into the model can largely increase the computation burden, and makes the model require significant overhead processing time during real-world applications.    

In contrast to existing literature, in this work, the proposed SAPI model does not exhaustively exploit available information, nor does it making rigid assumptions on vehicle motion. It uses an abstract way to represent surrounding driving environment, and focuses on learning patterns from the environment representation and history trajectory.   

\section{Methodology}
\noindent In this section we present the detailed methodology of SAPI. Firstly we provide the statement of problem defined in this work, and then propose the environment representation strategy, followed by the detailed model architecture. Finally, we show the overall prediction and relevant model settings. 
\subsection{Problem statement}
In this work we investigate the problem of vehicle trajectory prediction at intersections, with the information of real-time high-definition (HD) maps, surrounding vehicle dynamics and history trajectory of a target vehicle. The proposed model encodes surrounding environment information and history positions of the target vehicle in past several seconds, learns motion patterns, and predicts the future trajectory of it. The problem can be formulated as follows: Assume $S_t =(S_{tx}, S_{ty})$, $S_{tx} \times S_{ty} \in \mathbb{R}^2$, is the position of the target vehicle at time $t$, and $E_t$ is the representation of surrounding environment information at time $t$. With the history length of $m$ time steps, the observed history sequence is $ \mathscr{O}_t =((S_{t-m+1}, E_{t-m+1}), (S_{t-m+2}, E_{t-m+2}),..., (S_{t}, E_{t}))$. The objective is to predict the future position of vehicles given desired prediction horizon $n$ based on $\mathscr{O}_t$, i.e., estimating $\mathscr{Q}_t = (S_{t+1},S_{t+2},...,S_{t+n})$. In this paper we adopts $m =12$ and $n=15$, with time gap between two adjacent time step being 0.4 seconds, which corresponds to  4.8-second history information, and predicting the trajectory in future 6 seconds.  

\subsection{Environment representation}
 Driving behavior can be affected by numerous factors in its surrounding environment, including road geometry, surrounding traffic, weather, right-of-way, etc. Such factors are not able to be fully observed. We use an abstract approach to represent two main types of surrounding environment information at each history time step: i) legally reachable areas (LRAs) of a target vehicle and; ii) surrounding vehicle dynamics. To maintain relative spatial information, single-channel images are used to encode such information.
Instead of directly drawing HD maps and surrounding vehicles on images similar to some existing work, an energy-based encoding is used to compute ``energy weights'' of each lane segment in the scene and also that of surrounding traffic. In a constructed single-channel image, we assume that the higher the pixel values, i.e., lighter colors, the lower the energy. 
Based on Lyapunov stability theorem, the system will have the intention to transit to states with lower energy \cite{plumbley1995lyapunov}, which reflects the intention of the target vehicle at a time step. 

\subsubsection{LRAs}
%motivation
Right-of-way is one of the most basic rules that a vehicle needs to follow when driving, especially at intersection regions. It indicates future movement options of a vehicle, e.g., going straight, turning left or turning right. Based on the HD maps used in this work, road structures are formed by different lane segments. We query real-time surrounding lane segments of a vehicle in the intersection region.  Based on right-of-way information,  lane segments are divided into two types based on whether a segment can be legally reached from the current position of target vehicle at a time step or not. The detailed strategy is shown as follows.
% definition
Given a vehicle at time step $t$, the LRAs $C$ of the vehicle is defined as follows: i) Current lane segment $c_1$ that the target vehicle locates in; ii) lane segments $c_2$ that fall in the range if searching $d$ meters forward from $c_1$; iii) neighbor lane segment $c_3$ if the vehicle is allowed to do left or right lane change at time $t$; iv) lane segments $c_4$ that fall in the range if searching $d$ meters forward from $c_3$. Note that both $c_2$, $c_3$ and $c_4$ can contain multiple lane segments.  $c_1$ and $c_2$ depict reachable areas if the vehicle continues driving forward in the future, while $c_3$ and $c_4$ show all potential regions the vehicle may reach in next few seconds if lane-change is allowed at time $t$. The LRAs at time $t$ is then encoded by a single-channel bird-eye view image $X_{1t}$, with the vehicle position be the mid-bottom of the image, depicting LRAs seen by the target vehicle at time $t$. When encoding, LRAs are assigned with pixel value $255$, and $0$ otherwise. Fig. \ref{fig:road} shows the illustration of the LRAs of driving environment in Fig. \ref{fig:env}, where white regions in the figure show LRAs at the investigated time step, depicting future movement options. Boxes in Fig. \ref{fig:env} are vehicles, and the greener the box is, the faster the vehicle drive.  
The mid-bottom of the image is the position of the target vehicle,  which is depicted as a dashed gray box in Fig. \ref{fig:road}. Note that the dashed box and lane segment region borders are only used to demonstrate the settings of LRAs, and are not included in the model input. 

Note that in this work traffic light information is not considered currently, however, such information can be easily integrated into the representation by utilizing right-of-way when available in the future.  
% advantages
Based on the representation, the surrounding map information of a given target vehicle can be encoded through an abstract strategy based on right-of-way. Though not considered in this work, the presentation provides the flexibility to include other environment factors affecting right-of-way when available, e.g., traffic light status, real-time road closure information, etc., making the model adaptive for future applications.

\subsubsection{Surrounding vehicle dynamics}
Surrounding traffic plays a significant role in affecting vehicles' future motion and intention \cite{ma2019trafficpredict}. For each history time step, another single channel bird-eye view image $X_{2t}$ is introduced to represent surrounding traffic dynamics of a target vehicle. We introduce an encoding strategy of surrounding vehicle dynamics based on their motion energy. Assume the mass of a surrounding vehicle $i$ in the scene is $m'_i$, and $m'_i$ is proportional to the its size $s_i$, i.e.,  $m'_i \propto s_i$.
With the velocity of the vehicle $v_{it}$ at time $t$, motion energy of $i$ is calculated as $E_{motion} = \frac{1}{2}m'_iv_{it}^2$, $s.t.$ $E_{motion} \propto s_i v_{it}^2$. Then the pixel value of surrounding NPCs in the image is calculated through the distribution below:
\begin{equation}
    pixel\_value  = 255*(1-e^{-\frac{1}{0.01s_iv_{it}^2+1}})
    \label{eq:dynamics_value}
\end{equation}
Based on Eq. \ref{eq:dynamics_value}, in the representation image, pixels occupied by vehicles with greater motion energy will be encoded with a lower value (darker color). When drawing the bird-eye view image, vehicles are presented as boxes, with the sizes depending on the real size of the corresponding vehicles. The target vehicle is located at the mid-bottom of the image, and surrounding vehicles are plotted based on the relative position to the target vehicle. Fig. \ref{fig:traffic} illustrates a representation example of surrounding traffic at the investigated time step of Fig. \ref{fig:env}, in which the target vehicle is located at mid-bottom of the image. 
%Each box is a vehicle, with the size proportional to their real size. 
Different colors show the motion energy difference, with the corresponding pixel values computed based on Eq. \ref{eq:dynamics_value}. 
Note that both road geometry (map) and surrounding vehicle positions are rotated based on the heading of the target vehicle at the corresponding time step, with the target vehicle heads upward.

\subsubsection{Environment representation as model input}

\begin{figure}
\begin{subfigure}{.24\textwidth}
  \centering
  \includegraphics[width=1\linewidth]{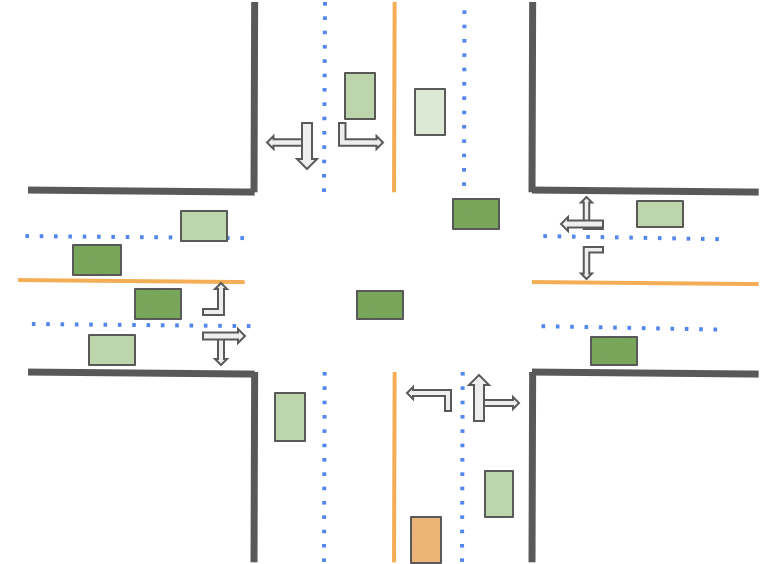}
  \caption{Surrounding environment}
  \label{fig:env}
\end{subfigure}
\begin{subfigure}{.24\textwidth}
  \centering
  \includegraphics[width=1\linewidth]{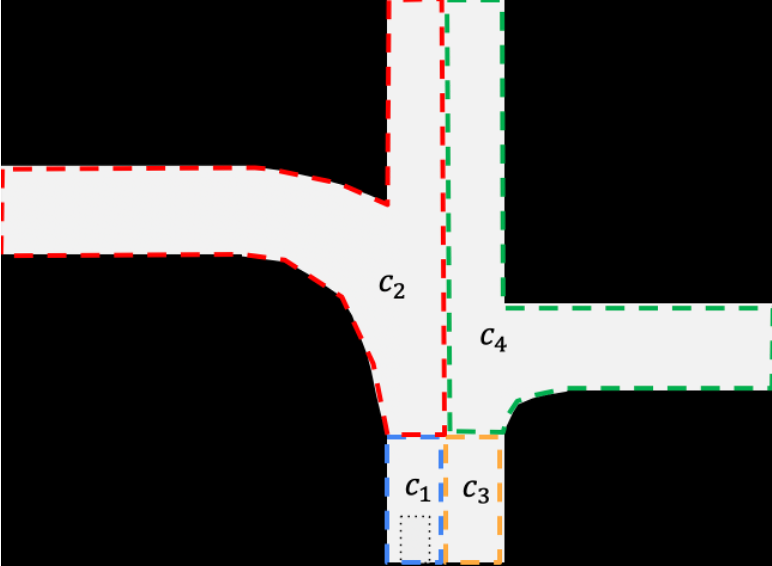}
  \caption{Legally reachable areas}
  \label{fig:road}
\end{subfigure}
\begin{subfigure}{.24\textwidth}
  \centering
  \includegraphics[width=1\linewidth]{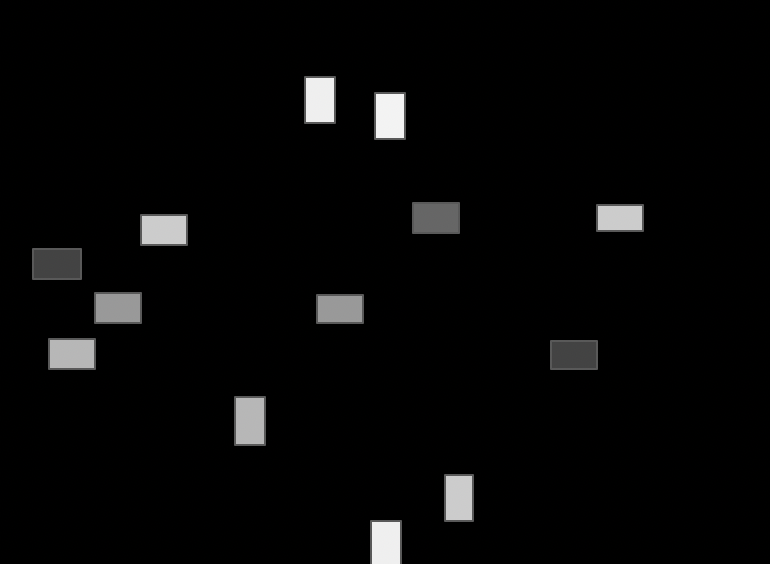}
  \caption{Surrounding vehicle dynamics}
  \label{fig:traffic}
\end{subfigure}
\begin{subfigure}{.24\textwidth}
  \centering
  \includegraphics[width=1\linewidth]{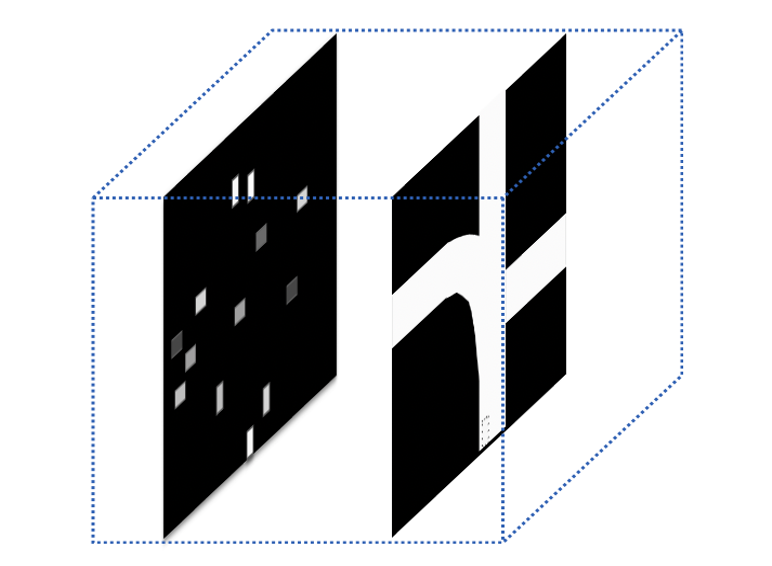}
  \caption{Environment representation}
  \label{fig:env_rep}
\end{subfigure}
\caption{Environment representation at a time step}
\label{fig:representation}
\end{figure}

\begin{figure*}
    \centering
    \includegraphics[scale=0.50]{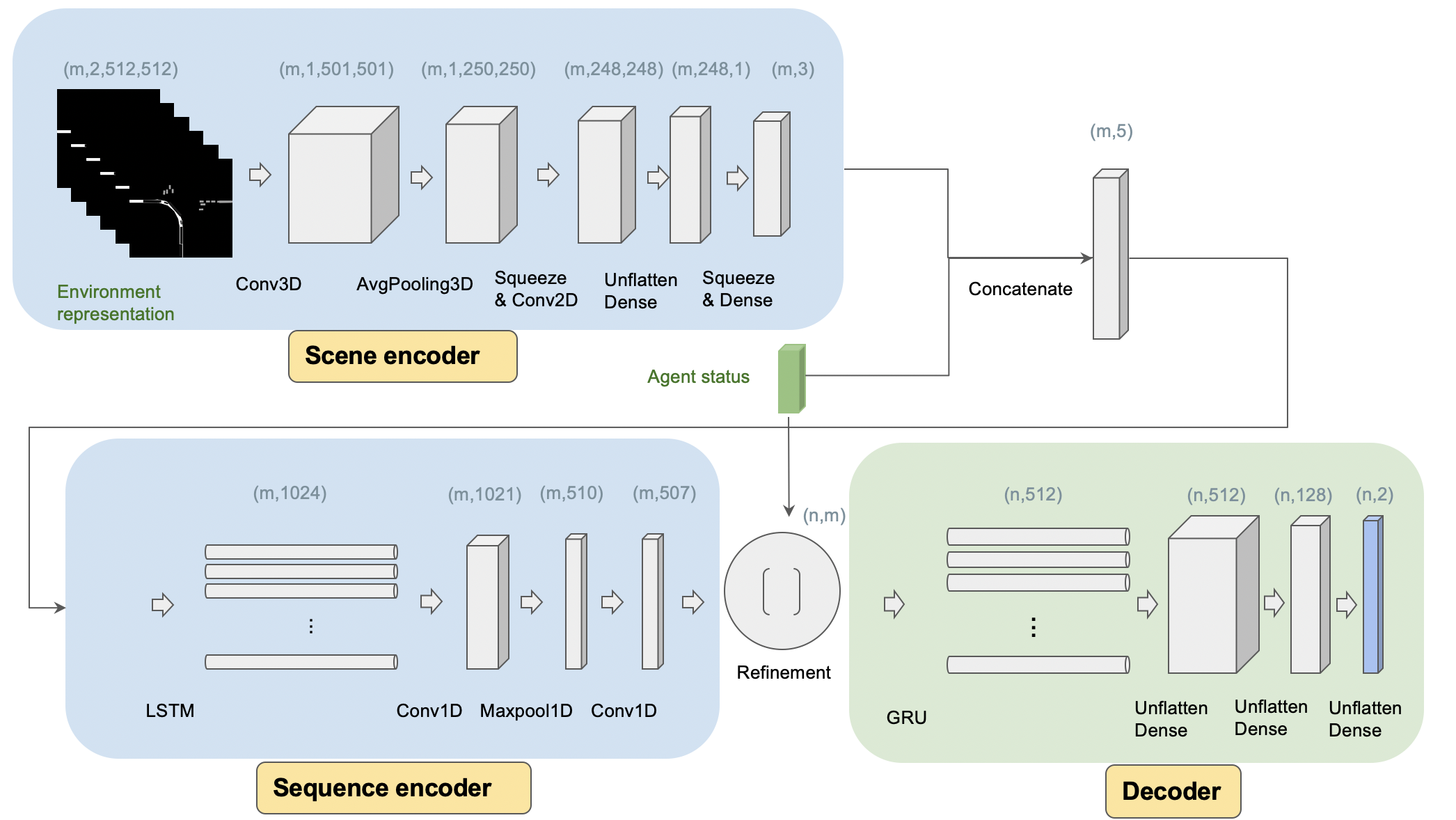}
    \caption{Proposed SAPI architecture}
    \label{fig:architecture}
\end{figure*}

Based on the aforementioned encoding strategy, by stacking $X_{1t}$ and $X_{2t}$, the surrounding environment at time $t$ of target vehicle is represented as a two-channel image as shown in Fig. \ref{fig:env_rep}, i.e., $E_t = stack(X_{1t},X_{2t})$. Along with the target vehicle position $S_t$, the input sequence to the model can be formed by concatenating $E_t$ and $S_t$ in past $m$ time steps.

\subsection{Model architecture}

The general idea behind SAPI is to learn patterns from both history trajectory and the surrounding environment of a given target vehicle, so as to accurately predict its future position. The proposed model architecture of SAPI is shown in Fig. \ref{fig:architecture}. The details of the architecture include:

\subsubsection{Scene encoder}
With $m$ history time steps, the environment representation sequence is firstly learned by a CNN-based scene encoder. First of all, a 3D convolutional layer is applied directly on top of two-channel images, features are extracted by a 3D average pooling strategy. Followed by a squeeze and 2D convolution layer, the model further learns patterns from the environment representation. After that two fully connected layers are constructed, and the environment information is then encoded. The output of scene encoder is a tensor $T_1$ with shape $(m,3)$.

 \subsubsection{Sequence encoder}
A sequence encoder is constructed to further learns patterns along with history trajectory. After learned by the scene encoder, the result environment encoding $T_1$ is concatenated with the history trajectory $S=(S_{t-m+1},S_{t-m+2},...,S_{t} )$, resulting a tensor $T_2$ with shape $(m,5)$. At the beginning, the sequence encoder learns sequence-related features in $T_2$ through an LSTM-based RNN, and passes it to a 1D convolutional layer. Followed by a maxpooling strategy and an additional 1D convolutional layer, patterns are learned into a feature tensor $T_3$. 

\subsubsection{Refiner}
History trajectory information reflects past driving patterns of vehicles. To fully utilize the history trajectory information, a refiner is defined in SAPI. By creating a short-cut between the raw history trajectory input and $T_3$, the actor refines the learned patterns by looking back to the raw history trajectory. The actor is defined in Eq. \ref{eq:refine}.
\begin{equation}
    T_4 = (W_1^T S + W_2^T T_3)^T
    \label{eq:refine}
\end{equation}
Where $W_1$ and $W_2$ are refinement weights subject to be learned.
\subsubsection{Decoder}
A decoder is then introduced to decode aforementioned learned patterns into a future trajectory. At the beginning, a GRU-based RNN is applied on refined features. Then two fully connected layers are constructed to further decoding features. Finally, the future trajectory is predicted with an additional dense layer. 

\subsection{Model setting and evaluation}
Based on the aforementioned model architecture, the proposed model has 5,000,329 parameters in total. To better measure sequence-related errors during training, we use the loss function as shown in 
Eq. \ref{eq:loss}, which is a Huber loss. 
\begin{equation}
    Loss = \sum_{i=1}^{n} \begin{cases} \frac{1}{2r} ({g}_i - {p}_i)^2 &
                           \text{ if } |{g}_i - {p}_i| < {r} \\
                           |{g}_i - {p}_i| - r &
                           \text{ otherwise }
                           \end{cases}
    \label{eq:loss}
\end{equation}
where $r$ is a pre-defined threshold, $g_i$ and $p_i$ are the ground truth position and predicted position at $i$-th time step respectively, and $n$ is the prediction horizon.
We evaluate the performance of the model based on following metrics:
    i) ADE: The displacement error average on all time steps for 6-second prediction horizon on all samples in the test set; 
    ii) 6s FDE: The displacement error at the final time step;
  iii) 4s FDE: The displacement error at 4th second in the future, showing the performance on mid-range prediction;
    iv) standard deviation of displacement errors at 4th second (4s FDE std) and 6th second (6s FDE std) on the whole test set. 
We also compare the performance of the model with following approaches:
\begin{enumerate}
    \item Resnet \cite{he2016deep}: Resnet is classic but powerful deep learning model, demonstrating good and robust performance in different fields \cite{lu2018deep, jung2017resnet, zhang2021resnet}. It has been adopted directly, or as a backbone, to conduct vehicle trajectory prediction in lots of work, and achieved good performance \cite{zhang2021resnet,huang2021traffic,zhang2020stinet}. In this work, we adopt Resnet34(v2) as a benchmark. The model contains 34 convolutional layers. The environment representation is firstly learned by Resnet. Then features are concatenated with agent history trajectory. Finally, features are decoded by two fully connected layers. Note that this benchmark model contains 21,771,260 parameters in total, which is approximately four times of the proposed model; 
    
    \item Vanilla LSTM \cite{altche2017lstm}: LSTM is a popular method for learning patterns in time-series data. In recent years, some researchers use it for vehicle trajectory prediction \cite{altche2017lstm,kim2017probabilistic,breuer2019analysis}. The benchmark model used in this work is a vanilla LSTM model with 1024 hidden units, and learns patterns directly on agent history trajectory, and decoded by two fully connected layers; 
    \item SAPI without map information: We consider the SAPI without the representation of LRAs;
    \item SAPI without surrounding traffic: We consider the SAPI without surrounding vehicle dynamics information.
\end{enumerate}

\section{Experiments and Results}
\noindent We construct a proprietary dataset through our autonomous vehicles to evaluate the proposed model performance. The data was collected from different intersections in real-world driving environment in Arizona, USA, and these intersections have different types, including four-leg intersections, T-type intersections, signal controlled, non-signal controlled and so on. Vehicles are firstly detected by our autonomous vehicles, and then corresponding history trajectories are extracted alone with real-time environment information, including map information, through a highly aggregated on-board pipeline. The dataset contains 77,876 training samples, 25,962 validation samples, and 25,962 test samples, and $d=100m$ is adopted as searching distance when generating LRAs. The model is trained with 2 NVIDIA 2080-Ti GPU, and learning rate adopted in this work is 0.003, with loss threshold $r=3$. The result statistics are shown in TABLE \ref{tab:result}. 

\begin{table}[ht]
\centering
\caption{Model performance. The values show the ADE/FDE metrics in meters. }
\begin{tabular}{p{1.4cm}|p{0.8cm}|p{ 1.0cm}|p{0.8cm}|p{1.0cm}|p{0.8cm}} 
\hline
 Method& 4s FDE & 4s FDE std &6s FDE &6s FDE std & 6s ADE \\
 \hline
 Resnet & 3.86 &6.28&9.52 &12.29&4.68 \\ 
 \hline 
 Vanilla LSTM & 3.46 &4.62&7.23 &9.81&2.90\\
 \hline 
 SAPI w/out LRAs  & 3.14 &4.33&6.52&9.24&2.68 \\
 \hline 
 SAPI w/out traffic  & 2.55 &3.73&5.23 &8.07&2.19  \\ 
 \hline 
Proposed & \textbf{2.11} & \textbf{3.10}& \textbf{4.32}&  \textbf{6.98} &\textbf{1.84}  \\
  \hline
\end{tabular}
\label{tab:result}
\end{table}
% statistics analysis here 

As can be seen from TABLE \ref{tab:result}, the proposed method outperforms all benchmarks in terms of ADE and FDE. For 6-second prediction, the proposed model has ADE/FDE of 1.84/4.32 meters. Considering the average vehicle length, which is around 4.2m, we claim that the proposed model demonstrates good performance. Furthermore, the result also indicates the promising performance of the proposed model from the perspective of model efficiency. Although the parameter size of Resnet34(v2) is about 4 times compared to that of the proposed model, the light-weighted SAPI still shows much better performance. It is ideal for real-world applications on autonomous vehicles, where computation time matters and computation resources are limited. It also demonstrates that the proposed model has higher prediction confidence, as it has lowest standard deviation for both mid-range 4-second prediction and the total 6-second prediction among all benchmark methods. 

\begin{figure}
    \centering
    \includegraphics[scale=0.5]{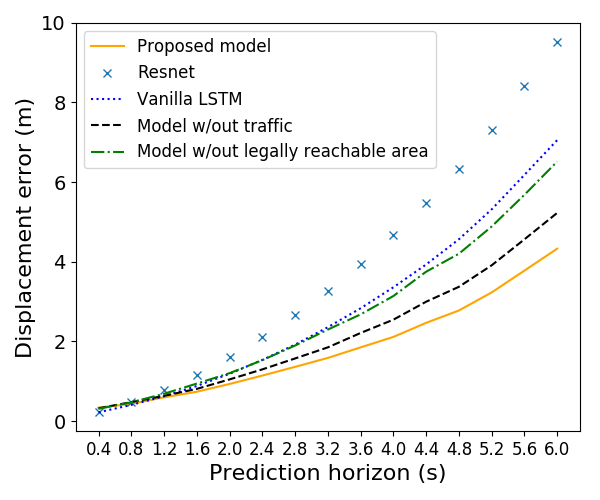}
    \caption{Displacement error at each prediction time step }
    \label{fig:horizon}
\end{figure}
The detailed displacement error at each prediction time step is shown in Fig. \ref{fig:horizon}. As demonstrated in Fig. \ref{fig:horizon}, as prediction horizon increases, the displacement error increases correspondingly for all investigated methods. However, the proposed model shows the best performance compared to benchmark methods in 6-second prediction, with smallest final displacement error and slowest displacement error increasing trend. Resnet and vanilla LSTM shows low prediction error at the beginning, but its prediction error increases rapidly when prediction horizon increases, demonstrating their limitations on long-horizon prediction. As depicted in the figure, model performance is impaired if LRAs or surrounding vehicle dynamics information is not available. The impact of LRAs to the model can be greater than that of surrounding traffic dynamics information, as the model shows worse performance if LRAs information is removed.   

\begin{figure}
\begin{subfigure}{.24\textwidth}
  \centering
  \includegraphics[width=1\linewidth]{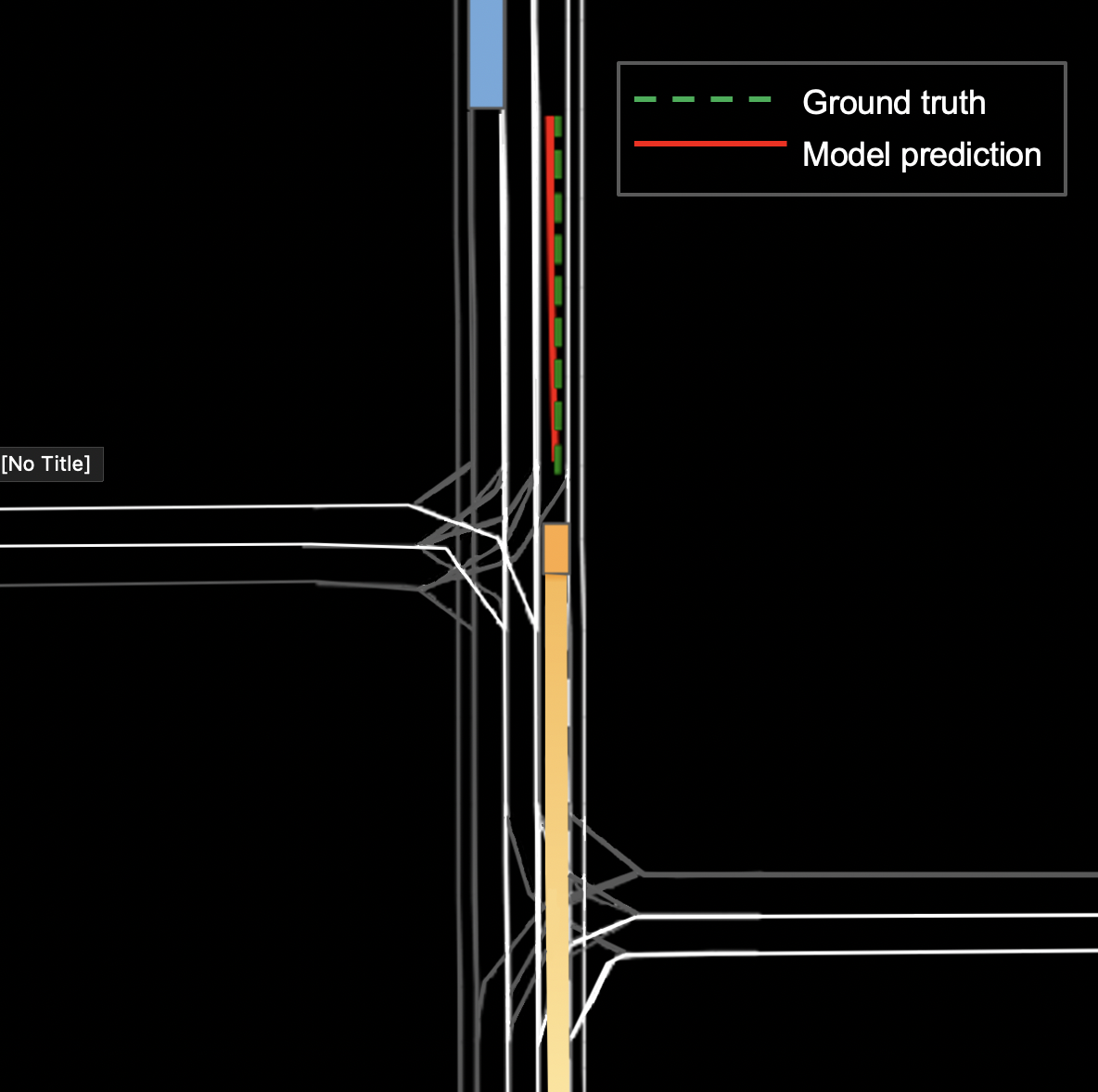}
  \caption{Driving straight example}
  \label{fig:straight}
\end{subfigure}
\begin{subfigure}{.24\textwidth}
  \centering
  \includegraphics[width=1\linewidth]{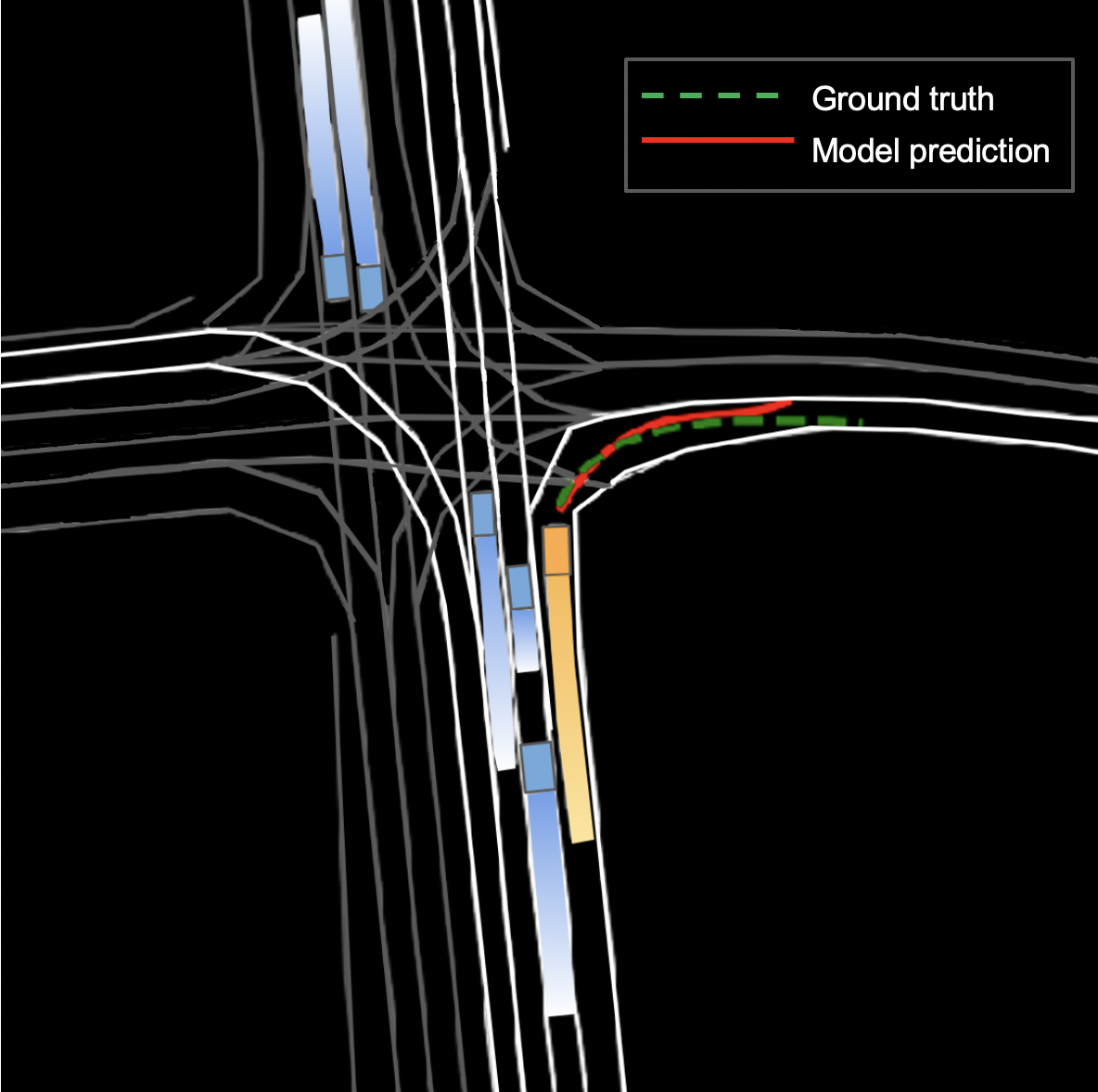}
  \caption{Turning example}
  \label{fig:turn}
\end{subfigure}
\begin{subfigure}{.24\textwidth}
  \centering
  \includegraphics[width=1\linewidth]{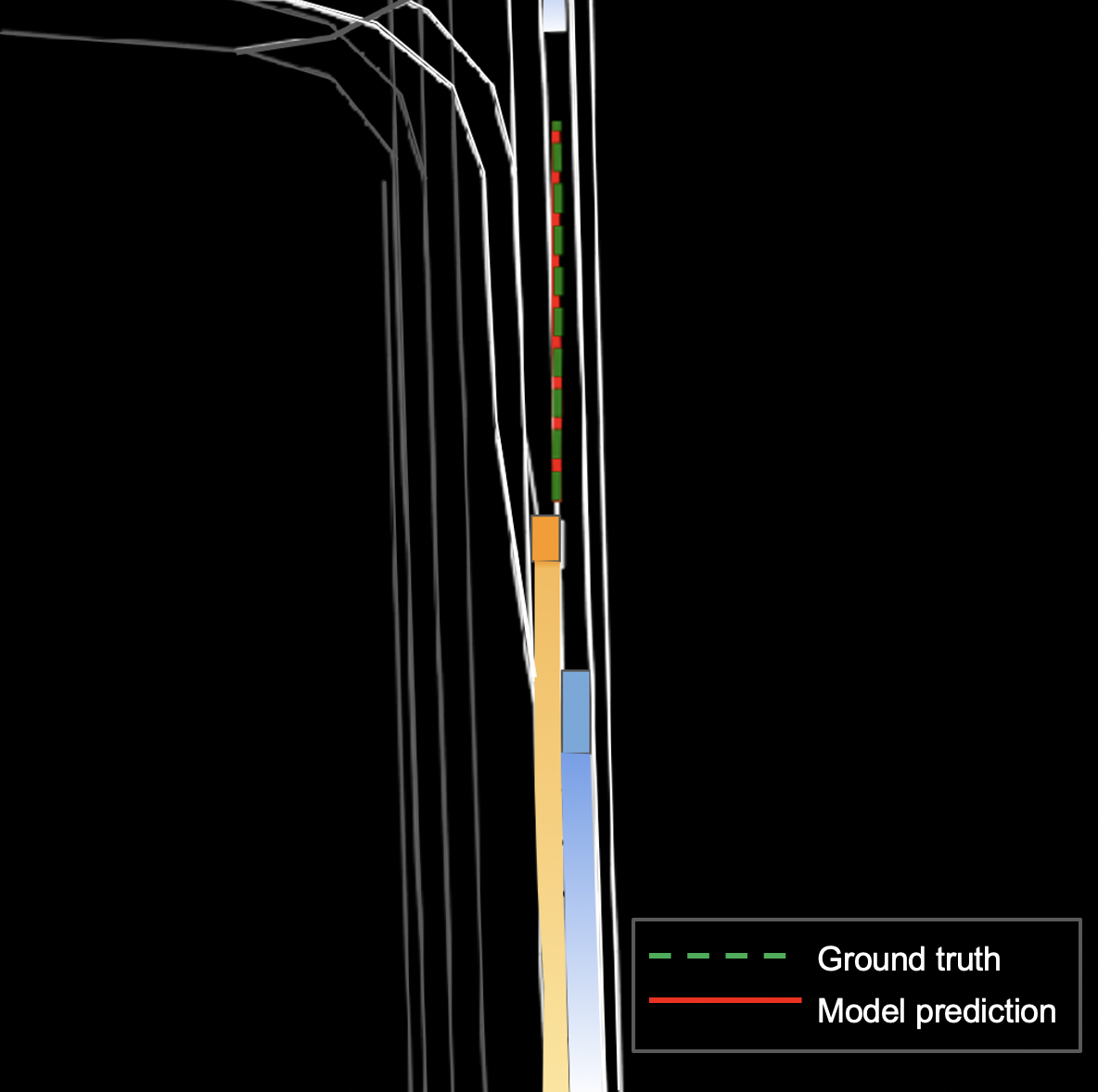}
  \caption{Lane change example}
  \label{fig:lane_change}
\end{subfigure}
\begin{subfigure}{.24\textwidth}
  \centering
  \includegraphics[width=1\linewidth]{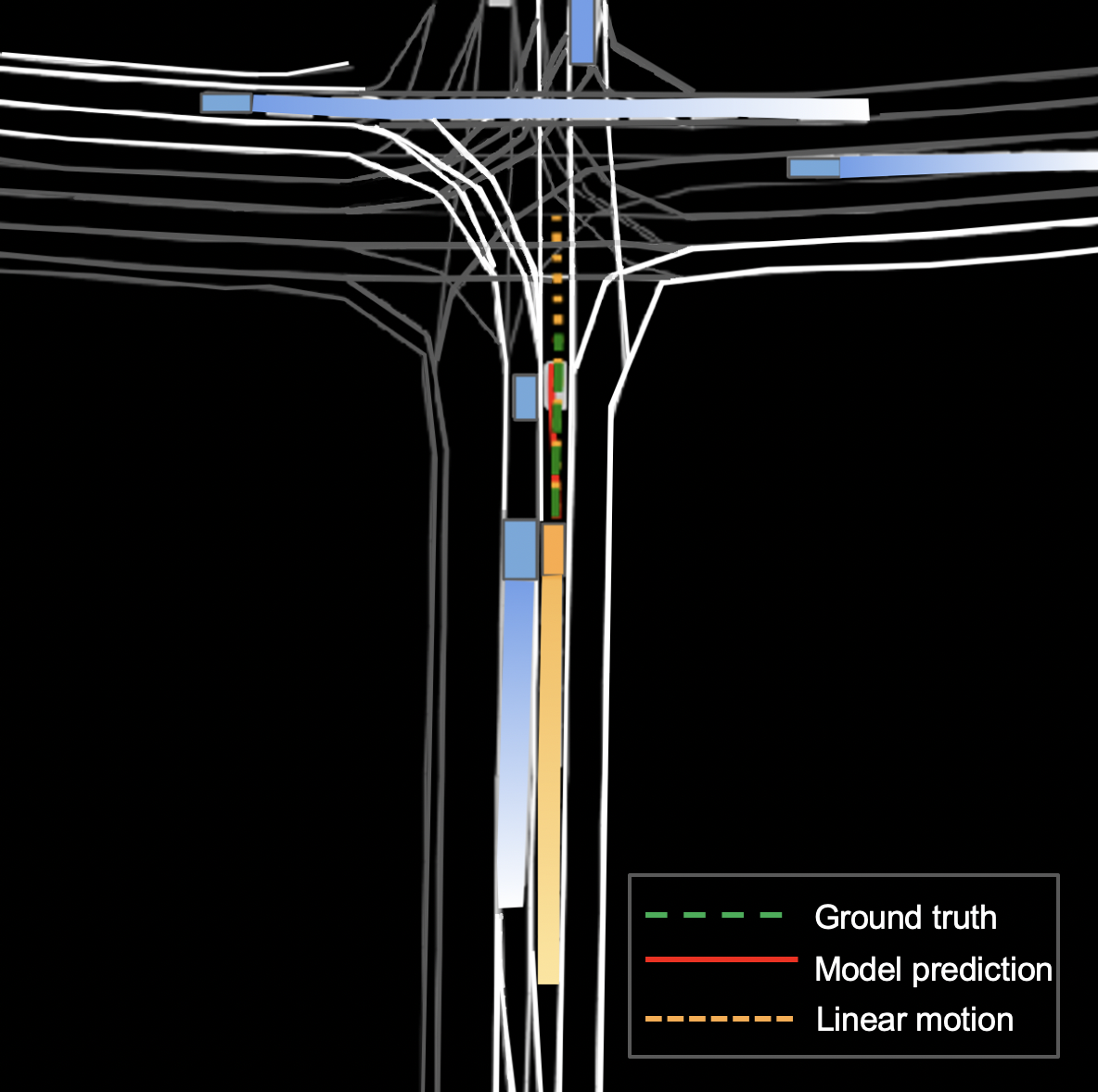}
  \caption{Stopping for traffic example}
  \label{fig:sft}
\end{subfigure}
\caption{Model result illustration on different scenarios.}
\label{fig:cases}
\end{figure}
Fig. \ref{fig:cases} illustrates detailed model performance on representative scenarios, including going straight, turning, lane-change, and stopping for traffic. For demonstration purpose, in Fig. \ref{fig:cases}, the center of each image is the latest position of target vehicle, which heads upward and is colored in yellow. In the figure, all latest positions of surrounding vehicles are shown in bounded blue boxes. History of vehicles in the scene are shown in faded color.  In the figure lane boundaries are colored in white if the lane is the same direction as the target vehicle, otherwise boundary lines are shown in gray. Virtual lanes (lanes without physical lane marks) are also included in the visualization.

As demonstrated in Fig. \ref{fig:cases}, the proposed SAPI model shows good performance in different scenarios. Fig. \ref{fig:straight} shows that the model can accurately predict the trajectory of target vehicles which is moving straight in the future. In the turning case shown in Fig. \ref{fig:turn}, the proposed model can accurately predict the right-turn motion of the target vehicle in advance. Though with a small final displacement error, the model still demonstrates high  performance in the scenario. Fig. \ref{fig:lane_change} showcases the SAPI model can successfully predict lane change behavior before it happens. As can be seen from  Fig. \ref{fig:sft}, the proposed model can also learn the impacts of surrounding traffic. For comparison, orange dashed line in Fig. \ref{fig:sft} is the result of driving without slowing down (linear motion). Although traffic light status is not included when training, the proposed model can still predict that target vehicle will slow down in the future for crossing traffic. Because there are surrounding vehicles drive from right to left, proposing conflicts with the driving direction of the target vehicle. 
All scenarios shown in Fig. \ref{fig:cases} are critic safety concerns in real-world applications for autonomous vehicles. As aforementioned, the proposed model can accurately predict vehicle trajectories in such scenarios, which demonstrates its robustness and good prediction performance.

\section{Conclusion}
%The main contributions of this work are: 
\noindent In this paper we propose a learning-based vehicle trajectory prediction model, i.e., SAPI. It uses an abstract way to represent and encode surrounding environment, by utilizing information from real-time map, right-of-way, and surrounding traffic. Due to the abstract representation setting, the environment representation strategy is flexible, and can be extend to future applications to incorporate more critic surrounding environment information, such as traffic light status and real-time changes of traffic management. SAPI contains two encoders and one decoder. A refiner is also proposed in the work to conduct a look-back operation, in order to make full use of history trajectory information. We evaluate SAPI based on a proprietary dataset collected in real-world intersections through autonomous vehicles. It is demonstrated that SAPI shows promising performance when predicting vehicle trajectories at intersections,  and outperforms benchmark methods in 6-second vehicle trajectory prediction, with ADE/FDE being 1.84/4.32m respectively. We also show that the proposed model demonstrates good performance when predicting vehicle trajectories in different scenarios, which ensures the safety of real-world applications on autonomous vehicles. 

% further improvement by introducing real-time traffic light status

% \section*{Appendix}

%%%%%%%%%%%%%%%%%%%%%%%%%%%%%%%%%%%%%%%%%%%%%%%%%%%%%%%%%%%%%%%%%%%%%%%%%%%%%%%%

\bibliographystyle{IEEEtranref}
\bibliography{reference}

\begin{thebibliography}{10}
\providecommand{\url}[1]{#1}
\csname url@samestyle\endcsname
\providecommand{\newblock}{\relax}
\providecommand{\bibinfo}[2]{#2}
\providecommand{\BIBentrySTDinterwordspacing}{\spaceskip=0pt\relax}
\providecommand{\BIBentryALTinterwordstretchfactor}{4}
\providecommand{\BIBentryALTinterwordspacing}{\spaceskip=\fontdimen2\font plus
\BIBentryALTinterwordstretchfactor\fontdimen3\font minus
  \fontdimen4\font\relax}
\providecommand{\BIBforeignlanguage}[2]{{%
\expandafter\ifx\csname l@#1\endcsname\relax
\typeout{** WARNING: IEEEtran.bst: No hyphenation pattern has been}%
\typeout{** loaded for the language `#1'. Using the pattern for}%
\typeout{** the default language instead.}%
\else
\language=\csname l@#1\endcsname
\fi
#2}}
\providecommand{\BIBdecl}{\relax}
\BIBdecl

\bibitem{tang2018single}
Z.~Tang, G.~Wang, H.~Xiao, A.~Zheng, and J.-N. Hwang, ``Single-camera and
  inter-camera vehicle tracking and 3d speed estimation based on fusion of
  visual and semantic features,'' in \emph{Proceedings of the IEEE conference
  on computer vision and pattern recognition workshops}, 2018, pp. 108--115.

\bibitem{xiao2018group}
H.~Xiao, W.~Lin, B.~Sheng, K.~Lu, J.~Yan, J.~Wang, E.~Ding, Y.~Zhang, and
  H.~Xiong, ``Group re-identification: Leveraging and integrating multi-grain
  information,'' in \emph{Proceedings of the 26th ACM international conference
  on Multimedia}, 2018, pp. 192--200.

\bibitem{lin2019group}
W.~Lin, Y.~Li, H.~Xiao, J.~See, J.~Zou, H.~Xiong, J.~Wang, and T.~Mei, ``Group
  reidentification with multigrained matching and integration,'' \emph{IEEE
  transactions on cybernetics}, vol.~51, no.~3, pp. 1478--1492, 2019.

\bibitem{he2020ust}
H.~He, H.~Dai, and N.~Wang, ``Ust: Unifying spatio-temporal context for
  trajectory prediction in autonomous driving,'' in \emph{2020 IEEE/RSJ
  International Conference on Intelligent Robots and Systems (IROS)}.\hskip 1em
  plus 0.5em minus 0.4em\relax IEEE, 2020, pp. 5962--5969.

\bibitem{deo2018convolutional}
N.~Deo and M.~M. Trivedi, ``Convolutional social pooling for vehicle trajectory
  prediction,'' in \emph{Proceedings of the IEEE Conference on Computer Vision
  and Pattern Recognition Workshops}, 2018, pp. 1468--1476.

\bibitem{ding2019online}
W.~Ding and S.~Shen, ``Online vehicle trajectory prediction using policy
  anticipation network and optimization-based context reasoning,'' in
  \emph{2019 International Conference on Robotics and Automation (ICRA)}.\hskip
  1em plus 0.5em minus 0.4em\relax IEEE, 2019, pp. 9610--9616.

\bibitem{roy2019vehicle}
D.~Roy, T.~Ishizaka, C.~K. Mohan, and A.~Fukuda, ``Vehicle trajectory
  prediction at intersections using interaction based generative adversarial
  networks,'' in \emph{2019 IEEE Intelligent Transportation Systems Conference
  (ITSC)}.\hskip 1em plus 0.5em minus 0.4em\relax IEEE, 2019, pp. 2318--2323.

\bibitem{gan2023mgtr}
Y.~Gan, H.~Xiao, Y.~Zhao, E.~Zhang, Z.~Huang, X.~Ye, and L.~Ge, ``Mgtr:
  Multi-granular transformer for motion prediction with lidar,'' \emph{arXiv
  preprint arXiv:2312.02409}, 2023.

\bibitem{woo2017lane}
H.~Woo, Y.~Ji, H.~Kono, Y.~Tamura, Y.~Kuroda, T.~Sugano, Y.~Yamamoto,
  A.~Yamashita, and H.~Asama, ``Lane-change detection based on
  vehicle-trajectory prediction,'' \emph{IEEE Robotics and Automation Letters},
  vol.~2, no.~2, pp. 1109--1116, 2017.

\bibitem{ammoun2009real}
S.~Ammoun and F.~Nashashibi, ``Real time trajectory prediction for collision
  risk estimation between vehicles,'' in \emph{2009 IEEE 5th International
  Conference on Intelligent Computer Communication and Processing}.\hskip 1em
  plus 0.5em minus 0.4em\relax IEEE, 2009, pp. 417--422.

\bibitem{park2018sequence}
S.~H. Park, B.~Kim, C.~M. Kang, C.~C. Chung, and J.~W. Choi,
  ``Sequence-to-sequence prediction of vehicle trajectory via lstm
  encoder-decoder architecture,'' in \emph{2018 IEEE Intelligent Vehicles
  Symposium (IV)}.\hskip 1em plus 0.5em minus 0.4em\relax IEEE, 2018, pp.
  1672--1678.

\bibitem{wang2019trajectory}
Y.~Wang, Z.~Liu, Z.~Zuo, Z.~Li, L.~Wang, and X.~Luo, ``Trajectory planning and
  safety assessment of autonomous vehicles based on motion prediction and model
  predictive control,'' \emph{IEEE Transactions on Vehicular Technology},
  vol.~68, no.~9, pp. 8546--8556, 2019.

\bibitem{ji2016path}
J.~Ji, A.~Khajepour, W.~W. Melek, and Y.~Huang, ``Path planning and tracking
  for vehicle collision avoidance based on model predictive control with
  multiconstraints,'' \emph{IEEE Transactions on Vehicular Technology},
  vol.~66, no.~2, pp. 952--964, 2016.

\bibitem{fagnant2015preparing}
D.~J. Fagnant and K.~Kockelman, ``Preparing a nation for autonomous vehicles:
  opportunities, barriers and policy recommendations,'' \emph{Transportation
  Research Part A: Policy and Practice}, vol.~77, pp. 167--181, 2015.

\bibitem{kim2017probabilistic}
B.~Kim, C.~M. Kang, J.~Kim, S.~H. Lee, C.~C. Chung, and J.~W. Choi,
  ``Probabilistic vehicle trajectory prediction over occupancy grid map via
  recurrent neural network,'' in \emph{2017 IEEE 20th International Conference
  on Intelligent Transportation Systems (ITSC)}.\hskip 1em plus 0.5em minus
  0.4em\relax IEEE, 2017, pp. 399--404.

\bibitem{usdot_2018}
\BIBentryALTinterwordspacing
USDOT, ``Intersection safety,'' Jul 2018. [Online]. Available:
  \url{https://highways.dot.gov/research-programs/safety/intersection-safety}
\BIBentrySTDinterwordspacing

\bibitem{cui2019multimodal}
H.~Cui, V.~Radosavljevic, F.-C. Chou, T.-H. Lin, T.~Nguyen, T.-K. Huang,
  J.~Schneider, and N.~Djuric, ``Multimodal trajectory predictions for
  autonomous driving using deep convolutional networks,'' in \emph{2019
  International Conference on Robotics and Automation (ICRA)}.\hskip 1em plus
  0.5em minus 0.4em\relax IEEE, 2019, pp. 2090--2096.

\bibitem{hong2019rules}
J.~Hong, B.~Sapp, and J.~Philbin, ``Rules of the road: Predicting driving
  behavior with a convolutional model of semantic interactions,'' in
  \emph{Proceedings of the IEEE/CVF Conference on Computer Vision and Pattern
  Recognition}, 2019, pp. 8454--8462.

\bibitem{messaoud2019non}
K.~Messaoud, I.~Yahiaoui, A.~Verroust-Blondet, and F.~Nashashibi, ``Non-local
  social pooling for vehicle trajectory prediction,'' in \emph{2019 IEEE
  Intelligent Vehicles Symposium (IV)}.\hskip 1em plus 0.5em minus 0.4em\relax
  IEEE, 2019, pp. 975--980.

\bibitem{houenou2013vehicle}
A.~Houenou, P.~Bonnifait, V.~Cherfaoui, and W.~Yao, ``Vehicle trajectory
  prediction based on motion model and maneuver recognition,'' in \emph{2013
  IEEE/RSJ international conference on intelligent robots and systems}.\hskip
  1em plus 0.5em minus 0.4em\relax IEEE, 2013, pp. 4363--4369.

\bibitem{schubert2008comparison}
R.~Schubert, E.~Richter, and G.~Wanielik, ``Comparison and evaluation of
  advanced motion models for vehicle tracking,'' in \emph{2008 11th
  international conference on information fusion}.\hskip 1em plus 0.5em minus
  0.4em\relax IEEE, 2008, pp. 1--6.

\bibitem{yi2015vehicle}
S.~G. Yi, C.~M. Kang, S.-H. Lee, and C.~C. Chung, ``Vehicle trajectory
  prediction for adaptive cruise control,'' in \emph{2015 IEEE Intelligent
  Vehicles Symposium (IV)}.\hskip 1em plus 0.5em minus 0.4em\relax IEEE, 2015,
  pp. 59--64.

\bibitem{barrios2015trajectory}
C.~Barrios, Y.~Motai, and D.~Huston, ``Trajectory estimations using
  smartphones,'' \emph{IEEE Transactions on Industrial Electronics}, vol.~62,
  no.~12, pp. 7901--7910, 2015.

\bibitem{xie2017vehicle}
G.~Xie, H.~Gao, L.~Qian, B.~Huang, K.~Li, and J.~Wang, ``Vehicle trajectory
  prediction by integrating physics-and maneuver-based approaches using
  interactive multiple models,'' \emph{IEEE Transactions on Industrial
  Electronics}, vol.~65, no.~7, pp. 5999--6008, 2017.

\bibitem{gupta2018social}
A.~Gupta, J.~Johnson, L.~Fei-Fei, S.~Savarese, and A.~Alahi, ``Social gan:
  Socially acceptable trajectories with generative adversarial networks,'' in
  \emph{Proceedings of the IEEE Conference on Computer Vision and Pattern
  Recognition}, 2018, pp. 2255--2264.

\bibitem{plumbley1995lyapunov}
M.~D. Plumbley, ``Lyapunov functions for convergence of principal component
  algorithms,'' \emph{Neural Networks}, vol.~8, no.~1, pp. 11--23, 1995.

\bibitem{ma2019trafficpredict}
Y.~Ma, X.~Zhu, S.~Zhang, R.~Yang, W.~Wang, and D.~Manocha, ``Trafficpredict:
  Trajectory prediction for heterogeneous traffic-agents,'' in
  \emph{Proceedings of the AAAI Conference on Artificial Intelligence},
  vol.~33, no.~01, 2019, pp. 6120--6127.

\bibitem{he2016deep}
K.~He, X.~Zhang, S.~Ren, and J.~Sun, ``Deep residual learning for image
  recognition,'' in \emph{Proceedings of the IEEE conference on computer vision
  and pattern recognition}, 2016, pp. 770--778.

\bibitem{lu2018deep}
Z.~Lu, X.~Jiang, and A.~Kot, ``Deep coupled resnet for low-resolution face
  recognition,'' \emph{IEEE Signal Processing Letters}, vol.~25, no.~4, pp.
  526--530, 2018.

\bibitem{jung2017resnet}
H.~Jung, M.-K. Choi, J.~Jung, J.-H. Lee, S.~Kwon, and W.~Young~Jung,
  ``Resnet-based vehicle classification and localization in traffic
  surveillance systems,'' in \emph{Proceedings of the IEEE conference on
  computer vision and pattern recognition workshops}, 2017, pp. 61--67.

\bibitem{zhang2021resnet}
Z.~Zhang, ``Resnet-based model for autonomous vehicles trajectory prediction,''
  in \emph{2021 IEEE International Conference on Consumer Electronics and
  Computer Engineering (ICCECE)}.\hskip 1em plus 0.5em minus 0.4em\relax IEEE,
  2021, pp. 565--568.

\bibitem{huang2021traffic}
K.~Huang, ``Traffic agent movement prediction using resnet-based model,'' in
  \emph{2021 6th International Conference on Intelligent Computing and Signal
  Processing (ICSP)}.\hskip 1em plus 0.5em minus 0.4em\relax IEEE, 2021, pp.
  136--139.

\bibitem{zhang2020stinet}
Z.~Zhang, J.~Gao, J.~Mao, Y.~Liu, D.~Anguelov, and C.~Li, ``Stinet:
  Spatio-temporal-interactive network for pedestrian detection and trajectory
  prediction,'' in \emph{Proceedings of the IEEE/CVF Conference on Computer
  Vision and Pattern Recognition}, 2020, pp. 11\,346--11\,355.

\bibitem{altche2017lstm}
F.~Altch{\'e} and A.~de~La~Fortelle, ``An lstm network for highway trajectory
  prediction,'' in \emph{2017 IEEE 20th International Conference on Intelligent
  Transportation Systems (ITSC)}.\hskip 1em plus 0.5em minus 0.4em\relax IEEE,
  2017, pp. 353--359.

\bibitem{breuer2019analysis}
A.~Breuer, S.~Elflein, T.~Joseph, J.-A. Bolte, S.~Homoceanu, and
  T.~Fingscheidt, ``Analysis of the effect of various input representations for
  lstm-based trajectory prediction,'' in \emph{2019 IEEE Intelligent
  Transportation Systems Conference (ITSC)}.\hskip 1em plus 0.5em minus
  0.4em\relax IEEE, 2019, pp. 2728--2735.

\end{thebibliography}

\end{document}